\documentclass[10pt,twocolumn,letterpaper]{article}

\PassOptionsToPackage{table}{xcolor}

\usepackage[pagenumbers]{cvpr}   

\usepackage{graphicx}
\usepackage{amsmath}
\usepackage{amssymb}
\usepackage{booktabs}
\usepackage{realboxes}
\usepackage[ruled]{algorithm2e}

\SetAlFnt{\small}
\SetAlCapFnt{\small}
\SetAlCapNameFnt{\small}
\SetAlCapHSkip{0pt}

\definecolor{cvprblue}{rgb}{0.21,0.49,0.74}
\usepackage[pagebackref,breaklinks,colorlinks,allcolors=cvprblue]{hyperref}

\usepackage[capitalize]{cleveref}
\usepackage{capt-of}   
\crefname{section}{Sec.}{Secs.}
\Crefname{section}{Section}{Sections}
\Crefname{table}{Table}{Tables}
\crefname{table}{Tab.}{Tabs.}

\newcommand{\methodname}{SeamGen}

\definecolor{valueColor}{RGB}{235, 245, 255}
\definecolor{gtColor}{RGB}{255, 245, 235}

\def\paperID{0000}
\def\confName{CVPR}
\def\confYear{2026}

\begin{document}

\title{\methodname: Artist-Aligned UV Seam Generation via Graph Flow Matching}

\author{%
  Hao Xu$^{1}$ \quad
  Yuqing Zhang$^{1}$ \quad
  Yiqian Wu$^{1}$ \quad
  Xueqi Ma$^{2}$ \quad
  Ding Liang$^{2}$ \quad
  Yan-Pei Cao$^{2}$
  \\[0.3em]
  Ying-Tian Liu$^{2}$ \quad
  Xiaogang Jin$^{1}$\thanks{Corresponding author.}
  \\[0.4em]
  $^{1}$State Key Lab of CAD\&CG, Zhejiang University \quad
  $^{2}$VAST
}

\twocolumn[{%
  \begin{@twocolumnfalse}
    \maketitle
    \begin{center}
      \centering
      \includegraphics[width=0.99\textwidth]{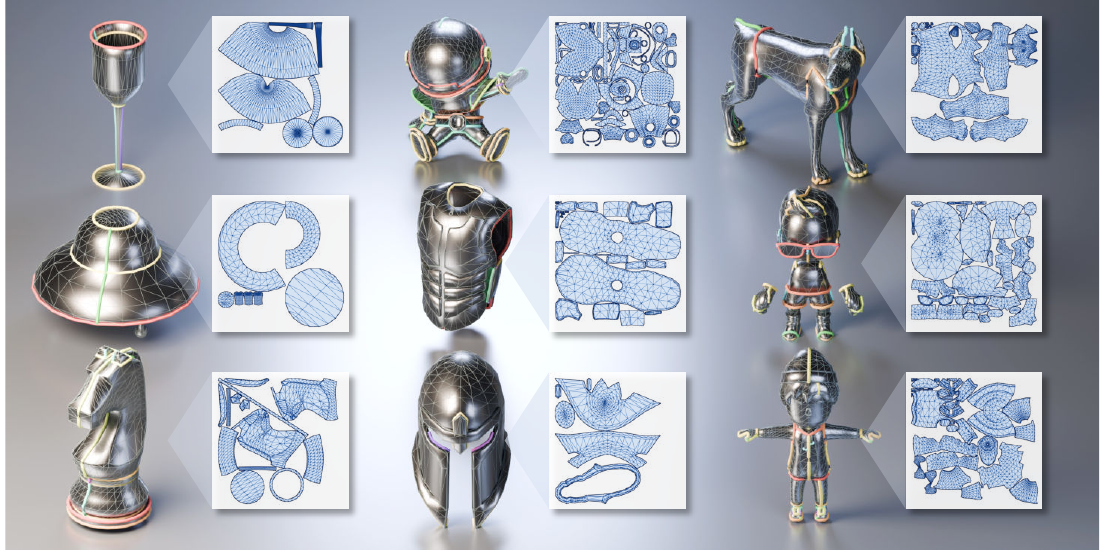}
      \captionof{figure}{%
        \textbf{\methodname{} generates UV seams that match artist conventions.}
        Instead of relying on hand-crafted distortion objectives for optimization, \methodname{} models the distribution of artist-authored seam layouts on irregular mesh graphs, generating coherent seam paths and well-structured UV islands that preserve semantic parts, reduce unnecessary fragmentation, place seams away from visually salient regions, and better support texture authoring in practice.%
      }\label{fig:teaser}
    \end{center}
    \vspace{1em}
  \end{@twocolumnfalse}
}]

\begin{abstract}
UV seam placement is a critical yet labor-intensive step in 3D content creation, requiring artists to balance chart shape, seam concealment, and alignment with semantic and geometric features. Existing automatic methods are primarily based on per-object optimization, relying on handcrafted objectives to avoid distortion or on proxies from pretrained models to inject semantic information. However, these strategies are not always well aligned with seams used in industrial production pipelines, often resulting in layouts that deviate from artist-preferred seam patterns and practical production requirements. 
To address these limitations, we propose \textit{\methodname{}}, a generative model for UV seam generation that aligns with artist preferences and production requirements. 
Instead of depending on manually designed objectives and constraints, \methodname{} learns the distribution of per-edge seam labels from a large corpus of existing seam layouts using a flow-matching generative model.
A key challenge is that typical Transformer architectures used in flow matching models are designed for sequential representations, such as point clouds, and cannot naturally account for mesh topology. To enable mesh-native learning, we design a Mesh Transformer backbone that interleaves local graph attention over mesh edges with global self-attention across vertices, capturing both fine-grained geometric cues and long-range topological coherence. 
To further improve inference-time controllability and quality, we exploit the training-free inpainting capability of flow models for both localized seam refinement and constraint-guided seam generation.
Extensive experiments show that by learning priors from professional seam layout data, \methodname{} produces UV layouts that better align with artist-authored preferences and achieve superior perceptual quality compared with distortion-based and semantic-proxy baselines.
\end{abstract}

\section{Introduction}
UV unwrapping is an essential step in 3D asset production, enabling textures and material details to be applied through a 2D parameterization of the mesh surface. Its quality depends critically on seam placement, i.e., where the surface is cut before flattening.  
Manual seam placement remains tedious and expertise-demanding. Desirable seams depend not only on geometric distortion, but also on perceptual visibility, semantic part boundaries, and artist conventions~\cite{DBLP:conf/visualization/ShefferH02,DBLP:conf/siggrapha/LucquinDB17,wang2025partuv}. To reduce this manual effort, 
automatic UV unwrapping methods seek to generate seams and parameterizations automatically, typically by formulating UV unwrapping as a geometry-driven optimization problem.

\begin{figure}
    \centering
    \vspace{-6mm}
    \includegraphics[width=\linewidth]{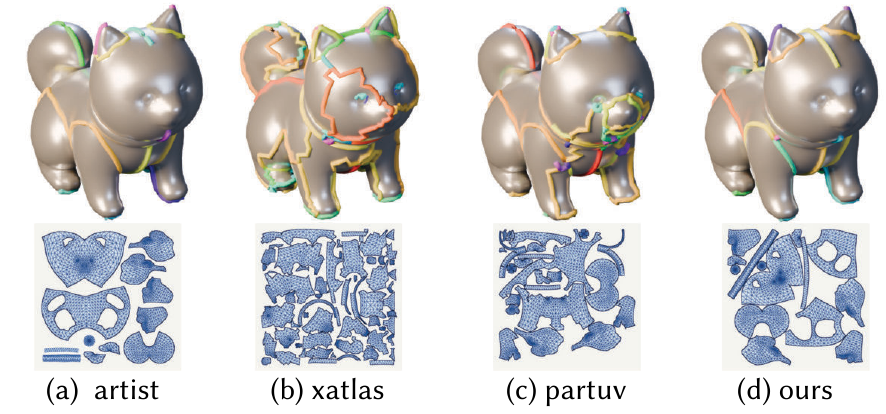}
    \caption{
    \textbf{Comparison with artist-authored seam placement.}
    (a) Artist-authored seams avoid salient regions (e.g., the face), favoring inconspicuous boundaries.
    (b-c) Distortion-oriented methods such as xatlas and PartUV produce seam layouts that differ substantially from the artist-authored reference, with seams often appearing in visually important facial regions.
    (d) \methodname{} better matches the artist-authored seam distribution, preserving the face while placing cuts along more natural boundaries.
    }
    \vspace{-6mm}
    \label{fig:artist}
\end{figure}

 
Most existing approaches are based on per-object optimization with geometric distortion objectives, such as angle or area preservation~\cite{levy2002least, sheffer2005abf, rabinovich2017scalable, li2018optcuts}, while more recent methods, such as PartUV~\cite{wang2025partuv}, additionally incorporate semantic part segmentation as a proxy for structured chart layouts. However, handcrafted objectives and semantic proxies are not always well aligned with the industrial production requirements. 
As a result, layouts produced by current methods often fail to place seams in ways that support practical texture authoring, such as avoiding visually salient regions and preserving texture continuity across important surface areas.
%
For the example of a dog model (see Fig. \ref{fig:artist}), artists typically avoid placing seams on the face. In contrast, methods such as xatlas~\cite{young2019xatlas} and PartUV~\cite{wang2025partuv} often introduce excessive seams in this region to reduce distortion, which disrupts texture continuity and increases the risk of visible artifacts.
This can be attributed to the gap between explicit optimization objectives and the complex seam design preferences encountered in real-world settings, where the latter often involve practical, perceptual, semantic, and geometric factors shaped by visual experience and production requirements.
Consequently, per-object optimization is inherently constrained. These observations motivate a data-driven formulation that learns seam placement directly from artist-authored examples.

To this end, we propose \textit{\methodname{}}, which formulates surface seam cutting as a conditional generation problem on the mesh graph, learning the distribution of artist-authored seam layouts conditioned on the input geometry.
Seam generation is challenging because seam edges occupy only a small fraction of all mesh edges, yet should still respect geometry, semantics, artist conventions, and topology on irregular mesh connectivity.  
Instead of independently identifying seams from all edges, we adopt flow matching~\cite{lipman2023flow} as our generative formulation, learning to transform random edge-wise noise into a coherent binary seam indicator on the mesh graph.
A key challenge is that typical Transformer architectures are designed for sequential modeling and are therefore better suited to representations such as point clouds, without naturally accounting for mesh topology. However, seam decisions depend on both global context, which coordinates cuts into coherent layouts across the entire shape, and local geometric cues, such as dihedral angles and curvature, which determine where seams should be placed in a topology-aware manner.
To enable mesh-native learning, we design a Mesh Transformer backbone that interleaves local graph attention over mesh edges with global self-attention across vertices, capturing both fine-grained geometric cues and long-range topological coherence.
%
To support localized control, we further exploit the learned flow model as a training-free conditional prior. 
Given sparse user constraints, such as seam-free regions or prescribed cut edges, \methodname{} clamps the specified edges during sampling and regenerates the remaining seam layout. 
The same mechanism also enables overlap self-refinement: when an initial layout produces invalid UV overlaps, we fix the reliable seams and resample only the problematic local region. 
This enables constrained editing and inference-time correction without additional training or a separate repair model.
Together, these designs enable \methodname{} to generate seam layouts that better match \textit{artist-authored priors} while remaining robust in practical UV unwrapping scenarios.

In summary, our contributions are as follows:
\begin{itemize}
    \item We formulate UV seam cutting as mesh-edge seam generation and propose \methodname{}, the first flow-matching framework that learns artist-authored seam layout priors from real-world UV data.
    \item We develop a mesh-native Mesh Transformer for seam prediction on mesh graphs, and reuse the learned flow model as a training-free conditional prior for constraint-guided completion and overlap self-refinement.
    \item We demonstrate that by learning priors from professional seam  data, \methodname{} produces better artist-aligned UV layouts with superior perceptual quality than baselines.
\end{itemize}

\section{Related Work}

\subsection{Geometry-driven UV Mapping}

Constructing a UV atlas requires determining both how to cut a surface into charts and how to flatten the resulting charts into a 2D domain. 
A long line of work addresses these problems through handcrafted geometric objectives~\cite{floater2005surface, sheffer2006mesh}. 
Classical surface parameterization methods include fixed-boundary formulations~\cite{tutte1963draw}, free-boundary and conformal approaches~\cite{levy2002least, sheffer2001parameterization, sheffer2005abf, mullen2008spectral, sawhney2021boundary, ben2009conformal}, and distortion-minimizing or injectivity-preserving optimization methods~\cite{sorkine2007arap, hormann2000mips, rabinovich2017scalable, li2018optcuts, pietroni2010almost, digne2016area, chen2012optimizing}.
These methods are effective for producing low-distortion and valid parameterizations, but they mainly optimize the geometric quality of the 2D map, such as angle preservation, area preservation, stretch minimization, or injectivity.
Because the choice of cuts strongly affects the parameterization domain and the structure of the resulting UV atlas, many geometry-driven methods also explicitly study seam placement and chart decomposition.
Bottom-up methods grow and merge charts under distortion bounds~\cite{julius2005d, DBLP:conf/visualization/SorkineCGL02, sander2003multi, carr2006rectangular}, forming the basis of practical tools such as xatlas~\cite{young2019xatlas} and Blender's Smart UV Project~\cite{blender}.
Other methods derive cuts from spectral analysis~\cite{zhou2004iso}, geometric feature lines~\cite{zhang2005feature, carlsson2014optimized}, or variational objectives that balance seam complexity and mapping distortion~\cite{liu2007mesh, poranne2017autocuts, sharp2018variational, li2018optcuts}.
However, they remain driven by predefined geometric criteria, rather than the semantic and perceptual preferences reflected in artist-authored UV layouts.

\subsection{Learning-based Surface Parameterization}

Recent work has explored neural formulations for surface parameterization and UV generation.
Neural parameterization methods learn continuous surface-to-domain mappings or distortion-aware representations~\cite{groueix2018papier, morreale2021neural, zhao2024flexpara, srinivasan2024nuvo, zhang2024flatten, liu2024dawand}.
Although these methods use neural networks, they are typically optimized with self-supervised geometric objectives, such as distortion, reconstruction, or injectivity-related losses.
Moreover, they commonly formulate UV generation as a point-wise mapping problem, where surface points or vertices are mapped continuously to 2D coordinates.
This differs from our setting, which focuses on predicting a discrete seam layout on the mesh connectivity and thereby determining a chart decomposition before flattening.
As a result, these methods mainly learn parameterizations or distortion-aware mappings, rather than modeling artist-authored seam distributions.

More recent methods move closer to data-driven UV decomposition.
PartUV~\cite{wang2025partuv} incorporates semantic part cues into chart decomposition by grouping faces according to both distortion and feature similarity.
Strips as Tokens~\cite{xu2026strips} introduces an autoregressive mesh generation framework with native UV segmentation, but its UV decomposition is generated as part of mesh generation rather than predicted for an existing input mesh.Concurrent work also explores data-driven seam generation.
SeamGPT~\cite{li2025seamgpt} and MeshTailor~\cite{ma2026meshtailor} learn seam layouts from data, while SeamCrafter~\cite{liu2025seamcrafter} further incorporates reinforcement learning to improve autoregressive seam generation.
Our method instead formulates seam cutting as generation on the mesh edge graph, using flow matching to jointly produce the full seam layout for a given input mesh.

\begin{figure*}[t]
    \centering
    \includegraphics[width=\linewidth]{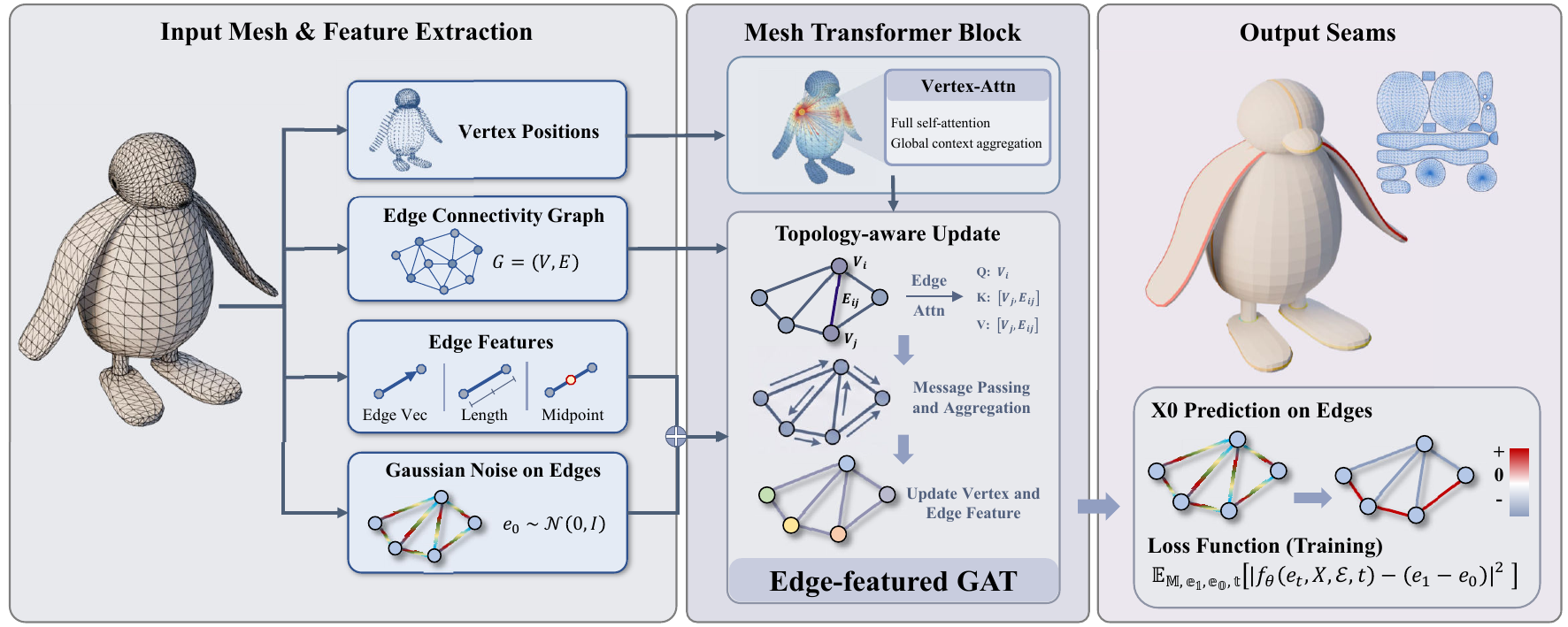}
\caption{
\textbf{Overview of the \methodname{} pipeline.}
Given an input mesh, \methodname{} casts UV seam cutting as generation on the mesh graph.
We build a mesh graph with geometric and topological features and parameterize the flow model with a Mesh Transformer.
Each block first applies global self-attention to position-aware vertex features, and then uses edge-featured graph attention to propagate local connectivity cues and update both vertex and edge features in a topology-aware manner.
Starting from random edge noise, the model generates a coherent per-edge seam mask, which is converted into seam cuts and used to unwrap the mesh into a UV atlas.
}
    \label{fig:pipeline}
\end{figure*}

\section{Method}


Given an input mesh, our method generates a per-edge seam layout, represented as a binary seam mask over mesh edges, which is then used to unwrap the mesh into a UV atlas.
We formulate this task as conditional seam generation on the mesh graph, where the model learns to capture the distribution of artist-authored seam layouts conditioned on the input mesh.
To this end, we instantiate a flow matching model for seam generation (Sec.~\ref{sec:formulation}), parameterize it with a Mesh Transformer that captures both global and local mesh cues (Sec.~\ref{sec:backbone}), and exploit the learned flow model as a training-free conditional generative prior, supporting both constraint-guided seam completion from sparse user inputs and automatic self-refinement of problematic UV regions (Sec.~\ref{sec:inpainting}). The overall pipeline is illustrated in Fig.~\ref{fig:pipeline}.

\subsection{Seam Cutting as Generation}
\label{sec:formulation}

A high-quality UV seam layout is rarely the result of optimizing a small set of explicit objectives. In practice, artist-authored seams reflect a complex combination of geometric, perceptual, semantic, and practical considerations: seams are often placed along sharp creases, hidden in visually inconspicuous regions, aligned with semantic part boundaries, and arranged to support downstream texture painting and distortion control. Many of these preferences are experience-driven, making them difficult to formalize as a fixed collection of handcrafted rules. As a result, learning an artist-authored prior offers a more effective solution for seam placement.

Motivated by this observation, we formulate UV seam cutting as a conditional generative problem on the mesh graph, where the model learns the distribution of artist-authored seam layouts conditioned on the input mesh.
We represent each mesh by its mesh graph $\mathcal{M}=(\mathcal{V},\mathcal{E})$, where $\mathcal{V}\in\mathbb{R}^{N\times3}$ denotes the vertex positions and $\mathcal{E}$ denotes the edge set.
Given $\mathcal{M}$, we seek to learn a model distribution $p_\theta(e \mid \mathcal{M})$ that approximates the artist-authored data distribution $q_{\mathrm{data}}(e \mid \mathcal{M})$, where $e$ is a binary seam mask over mesh edges, with $e_i = 1$ indicating that edge $i$ is selected as a UV seam.
We realize this generative process with flow matching.
Starting from Gaussian noise defined on mesh edges, the model learns a conditional flow that transports samples toward artist-authored seam masks conditioned on the input mesh. 
During training, we sample an artist seam mask $e_1$ and an edge-wise Gaussian noise vector $e_0$, construct an interpolation state $e_t=(1-t)e_0+t e_1$, and train the network to predict the corresponding velocity from the noisy edge state, mesh geometry, connectivity, and time. 
At inference time, we integrate the learned flow from noise to obtain continuous edge states, which are then discretized into a binary seam mask.
In this way, our method does not rely on manually specified seam-placement rules or handcrafted distortion objectives. 
Instead, it learns the distributional regularities of artist-authored UV seams directly from data, enabling generated layouts to reflect the implicit geometric, semantic, and perceptual priors observed in professional seam design.

\subsection{Mesh Transformer Backbone}
\label{sec:backbone}
A key challenge in learning artist-authored seam layouts is that seam placement is simultaneously local and global. Locally, whether an edge should be cut is strongly influenced by nearby surface geometry, such as sharp creases, curvature changes, part junctions, and occluded regions. Globally, these edge-level decisions must be coordinated across the entire mesh to form continuous cut paths and partition the surface into well-shaped UV islands. To address this challenge, we parameterize the flow velocity field with a Mesh Transformer backbone that reasons over both mesh connectivity and long-range surface dependencies.

As illustrated in Fig.~\ref{fig:pipeline}, our backbone couples global self-attention with topology-aware local graph attention. We initialize each vertex feature using a frequency positional encoding of its 3D coordinate. 
For each mesh edge $(i,j)\in\mathcal{E}$, we construct a geometric edge descriptor from the endpoint coordinates, including the relative offset $p_j-p_i$, edge length, normalized direction, and midpoint position. 
This descriptor provides local geometric cues for the graph-attention stream.
We concatenate it with the current noisy seam variable $e_{t,ij}$ and project the result into the edge feature space. A sinusoidal timestep embedding is further added to the edge feature, allowing the network to condition its prediction on the current point along the flow trajectory.

Each backbone block contains two complementary attention modules. The global self-attention module updates vertex features over the entire mesh, allowing distant surface regions to exchange information directly. This provides long-range context for coordinating seam decisions across the shape. 
The local graph attention module then updates vertex and edge features along mesh connectivity, injecting topology-aware geometric information into the edge representation. 
For each vertex, the vertex feature to be updated serves as the query, while neighboring vertex features and their associated edge features are used to construct keys and values for attention-based aggregation. The resulting update allows each vertex to gather local geometric and topological evidence from its one-ring neighborhood. We then update each edge feature using the updated features of its two incident vertices together with its previous edge representation.

\begin{figure*}[t]
    \centering
    \includegraphics[width=\linewidth]{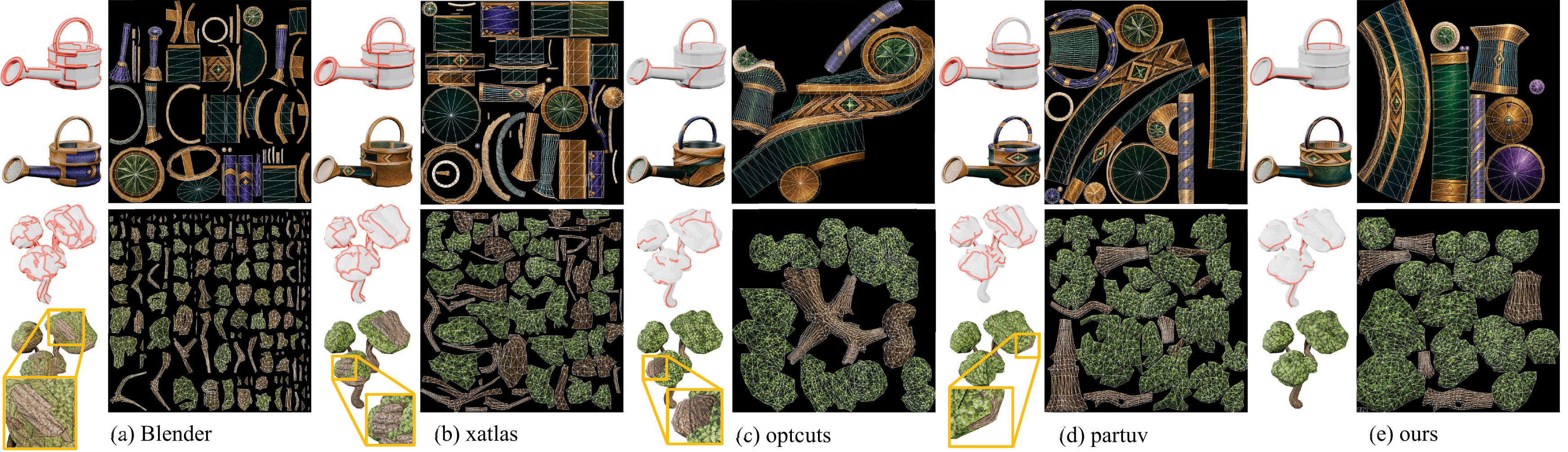} 
    \caption{\textbf{Comparison.} We compare our \methodname{} against Blender's Smart UV Project~\cite{blender}, xatlas~\cite{young2019xatlas}, OptCuts~\cite{li2018optcuts}, and PartUV~\cite{wang2025partuv}.
    For each case, we show renderings with seams overlaid (upper left), the results of semantic-aware UV space texture generation (bottom left), and the corresponding UV space textures (right). Problematic regions are highlighted with yellow zoom-in boxes.
    } 
    \label{fig:comp_tex}
    \vspace{-4mm}
\end{figure*}

By alternating global context aggregation and local topology-aware update, the backbone captures both the global organization of seam layouts and the local evidence that determines where seams should lie. Since mesh edges are undirected, we process each edge in both directions during message passing and average the two directional edge features before prediction. The final edge features are normalized and projected to a scalar output for each mesh edge, yielding the edge-wise velocity field used in our flow matching model.

\subsection{Conditional Seam Inpainting}
\label{sec:inpainting}
In production UV workflows, artists often refine an initial unwrap under partial constraints, such as protecting salient regions, enforcing selected seams, or revising charts that cause overlaps. 
We support this process with conditional seam inpainting, which reuses the learned flow model as a generative prior.

Following conditional diffusion sampling~\cite{lugmayr2022repaint}, we clamp the known edges throughout the sampling process and let the model predict the unknown ones.
Let $\mathcal{K} \subset \mathcal{E}$ denote the set of known edges with target seam labels $e_1^{\mathcal{K}} \in \{0,1\}^{|\mathcal{K}|}$.
During sampling, we follow the same procedure as in unconditional generation, but overwrite the values of known edges at each step using their corresponding flow interpolation:
\[
e_{t,i} = t\, e_{1,i}^{\mathcal{K}} + (1-t)\, e_{0,i}, \quad i \in \mathcal{K},
\]
where $e_0$ is the initial Gaussian noise.
The updated $e_t$ is then passed to the Mesh Transformer for the next denoising step.
Because our backbone propagates information through both global self-attention and local message passing, the clamped edges serve as contextual constraints for the remaining editable edges, enabling coherent conditional seam generation.

This conditional inpainting mechanism enables two practical applications.
First, it supports constraint-guided seam completion.
Given sparse user-provided constraints, such as seam-free regions or edges that should be cut, \methodname{} resamples the unconstrained edges conditioned on these inputs.
The generated layout respects the prescribed constraints while remaining consistent with the learned artist-authored seam prior.
This provides a convenient form of coarse controllability: users specify high-level editing intent, and the model completes the remaining seam layout through conditional sampling.
Second, the same mechanism enables automatic localized self-refinement.
Since overlap-freeness is not explicitly enforced by the generative objective, a small number of sampled layouts may produce self-overlapping UV islands after parameterization.
For these cases, we keep the reliable parts of the seam layout fixed and treat only the region around the problematic island as editable.
Conditional inpainting then locally resamples seams in the affected region while preserving the rest of the layout.
This reuses the pretrained model as a local seam prior without requiring a separate repair network, additional training, or a handcrafted post-processing objective.

\begin{table*}[t]
\definecolor{valColor}{RGB}{255, 245, 235}  
    \definecolor{wildColor}{RGB}{235, 245, 255} 
    \definecolor{distColor}{RGB}{240, 252, 239} 
    \definecolor{userColor}{RGB}{248, 240, 252} 
     \definecolor{1st}{HTML}{F8CBA6}
    \definecolor{2nd}{HTML}{FFE7CC}
    \centering
    \caption{\textbf{Quantitative comparisons.}
We evaluate seam generation performance using layout, distortion, runtime, and perceptual metrics. 
For layout statistics, we directly report the number of seam edges and UV islands, where values closer to the ground truth are better. 
We also report standard distortion metrics, runtime, and preference scores from both large vision-language models and human users.
\methodname{} achieves the closest layout statistics to the ground truth while obtaining the highest perceptual preference scores with comparable distortion and runtime.
\textcolor{1st}{$\blacksquare$} and \textcolor{2nd}{$\blacksquare$} and denote the 1st and 2nd places.
}
    \label{tab:structure}
    \small
    \setlength{\tabcolsep}{4.5pt} 
    \renewcommand{\arraystretch}{1.3}
    \resizebox{0.85\linewidth}{!}{
    \begin{tabular}{l cc ccc cc}
        \toprule
        Method
        & \textbf{VLM Score} $\uparrow$
        & \textbf{User Pref.} $\uparrow$ 
        & \textbf{\#Seam Edges $\sim GT$} 
        & \textbf{\#Islands $\sim GT$} 
        & \textbf{Ang. Distort.} $\sim 1$ 
        & \textbf{Area Distort.} $\sim 1$ 
        & \textbf{Running Time (s)} $\downarrow$ 
        \\
        \midrule
        
        Ground Truth 
        & -- 
        & -- 
        & 476.6  
        & 67.8  
        & 0.9110 
        & 2.4238  
        & -- 
        
        \\
        \midrule

        Blender 
        & 2.33  
        & 2.37 
        & 800.7  
        & 133.6 
        & 0.9207 
        & 1.1120 
        & \cellcolor{1st}{0.008} 
        
        \\
        
        xatlas 
        & 2.71 
        & 2.34 
        & 722.8  
        & 110.8  
        & \cellcolor{1st}{0.9817} 
        & \cellcolor{2nd}{1.0852} 
        & \cellcolor{2nd}{0.137}  
       
        \\
        
        optcuts 
        & 2.93  
        & 2.58 
        & 144.4  
        & 4.6  
        & 0.8845 
        & 1.1718 
        & 568.53  
        \\
        
        PartUV 
        & \cellcolor{2nd}{3.02} 
        & \cellcolor{2nd}{2.91}  
        & \cellcolor{2nd} 628.1  
        & \cellcolor{2nd} 96.1 
        & \cellcolor{2nd}{0.9749} 
        & \cellcolor{1st}{1.0843} 
        & 1.17 

        \\
        
        \methodname{} (Ours) 
        & \cellcolor{1st}{3.56} 
        & {\cellcolor{1st}{4.37}} 
        & {\cellcolor{1st} 471.5}  
        & {\cellcolor{1st}62.7}  
        & 0.9421 
        & 1.9087 
        & 2.87  
        \\
        
        \bottomrule
    \end{tabular}
    }
\end{table*}

\section{Experiments}
In this section, we first introduce the implementation details of \methodname{} (Sec.~\ref{sec:impl}), then present comparisons with state of the art methods (Sec.~\ref{sec:Comparison}), evaluate the effectiveness of each component through ablation studies (Sec.~\ref{sec:ablation}).We then demonstrate additional qualitative results, including symmetric seam layouts and component-wise UV generation for complex assets (Sec.~\ref{sec:additional_results}) and finally demonstrate inpainting applications (Sec.~\ref{sec:app_inpainting}).

\subsection{Implementation Details}
\label{sec:impl}
We collect artist-authored UV layouts from publicly available 3D assets, including meshes from Objaverse~\cite{objaverse,objaverseXL} and Sketchfab. 
We train the flow matching stage for 200 epochs on 8 NVIDIA A100 GPUs, using the AdamW optimizer with a fixed learning rate of $1\times10^{-5}$ and a batch size of 16.  
We use the $x$-prediction parameterization together with the $v$ loss for training. During training, each mesh is augmented with random rotations to improve generalization. At inference time, the final continuous seam predictions are binarized using a threshold of $\tau = 0.0$. We then use Blender's Python API to unwrap each mesh with the extracted seams using the angle-based unwrapping method. Detailed network architecture parameters, dataset collection and preprocessing procedures are provided in the supplementary material.

\subsection{Comparison}
\label{sec:Comparison}
We compare \methodname{} with automatic UV unwrapping baselines: Blender's Smart UV Project~\cite{blender}, xatlas~\cite{young2019xatlas}, OptCuts~\cite{li2018optcuts}, and PartUV~\cite{wang2025partuv}.

\subsubsection{Qualitative Comparison}
\label{sec:qualitative}

\begin{figure*}[p]
    \centering
    \includegraphics[width=\linewidth]{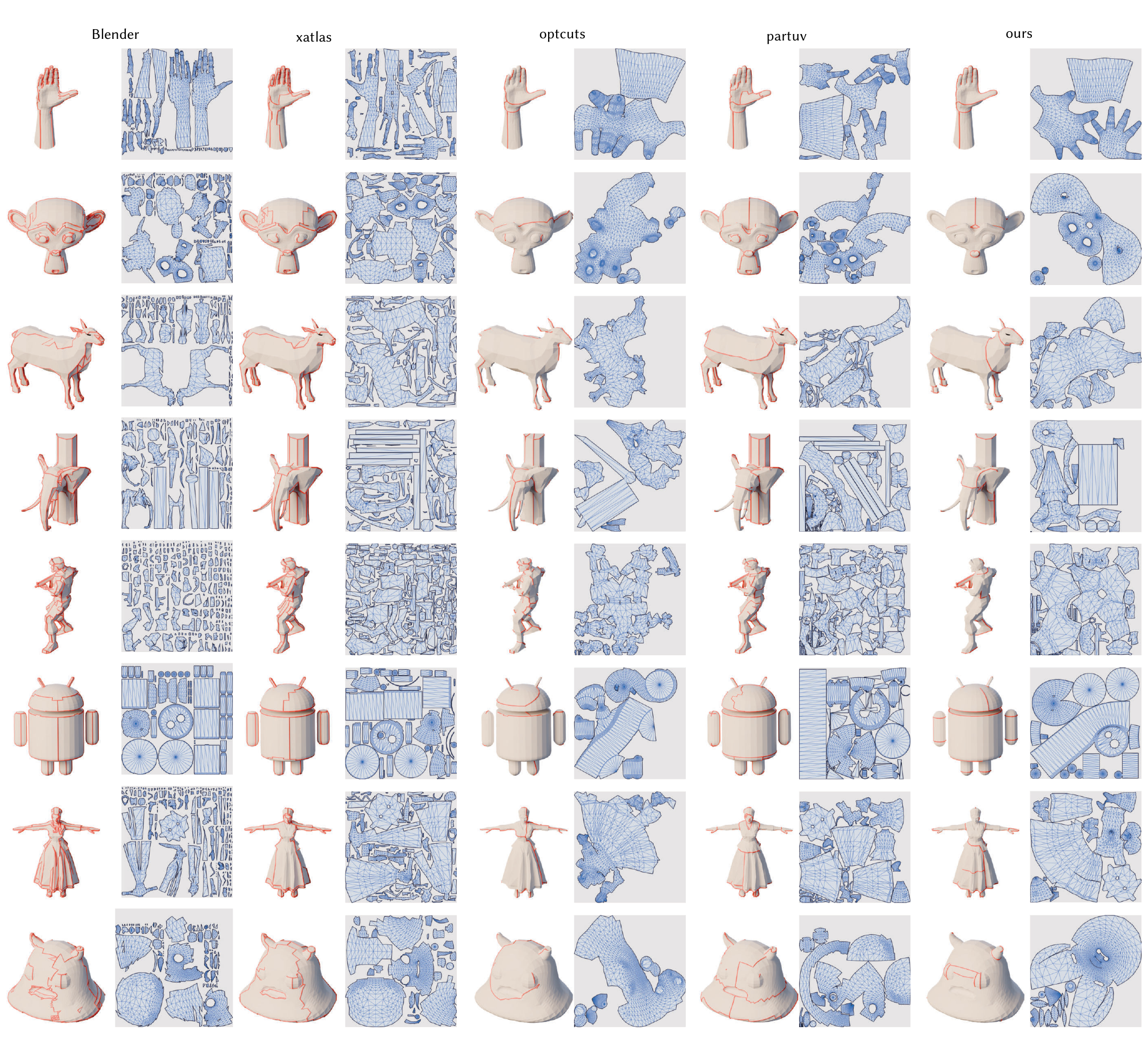}
    \caption{\textbf{Qualitative comparison} of UV unwrapping results on representative meshes. From left to right, we show the results of Blender~\cite{blender}, xatlas~\cite{young2019xatlas}, OptCuts~\cite{li2018optcuts}, PartUV~\cite{wang2025partuv}, and \methodname{} (ours). For each method, we visualize the seam layout on the mesh and its corresponding UV map; seam edges are highlighted in red. Compared with the baselines, \methodname{} produces more complete and coherent seam layouts, yielding UV atlases that more closely resemble artist-authored references.}
    \label{fig:qualitative}
\end{figure*}

We show qualitative comparisons in Fig.~\ref{fig:qualitative}. 
Blender~\cite{blender} and xatlas~\cite{young2019xatlas} often over-fragment the surface into many small UV islands, resulting in dense seam cuts that can break semantic coherence and pass through visually prominent regions. 
OptCuts~\cite{li2018optcuts} produces more consolidated charts, often mapping each connected surface region to a single large island, but the resulting island boundaries tend to be distorted and less aligned with semantic parts. 
PartUV~\cite{wang2025partuv} mitigates over-fragmentation by incorporating semantic part priors. 
However, its distortion-driven flattening process may still disrupt semantic continuity, resulting in seams that occasionally appear in visually important regions.
%
In contrast, \methodname{} produces more coherent seam layouts with fewer fragmented cuts and better-shaped UV islands. 
The generated seams are often aligned with natural geometric boundaries, such as creases, part junctions, and occluded regions, while avoiding visually prominent areas. 
This leads to more artist-friendly UV decompositions for downstream texture editing.

We further assess the semantic quality of UV layouts through their compatibility with 2D generative priors. Specifically, we use GPT-Image-2~\cite{openai2026gptimage2} to synthesize textures directly in UV space from position and normal maps (Fig.~\ref{fig:comp_tex}). Since 2D trained models naturally generate continuous textures within connected regions, the synthesis quality serves as a proxy for the semantic coherence and continuity of the UV layout.
Baselines with highly fragmented UV charts often destroy global semantic context, leading to severe color bleeding and disjointed patterns due to insufficient structural continuity across isolated islands.
In contrast, our approach preserves contiguous semantic regions, significantly mitigating these artifacts and producing highly coherent textures.

\subsubsection{Quantitative Comparison}
\label{sec:quantitative}
We use a held-out validation set of 200 meshes for evaluation. 
OptCuts only successfully processes 170 out of the 200 validation meshes due to its stricter input requirements; we therefore report its statistics on these valid cases. All other methods are evaluated on the full validation set.

\begin{figure}[t]
    \centering
    \includegraphics[width=\linewidth]{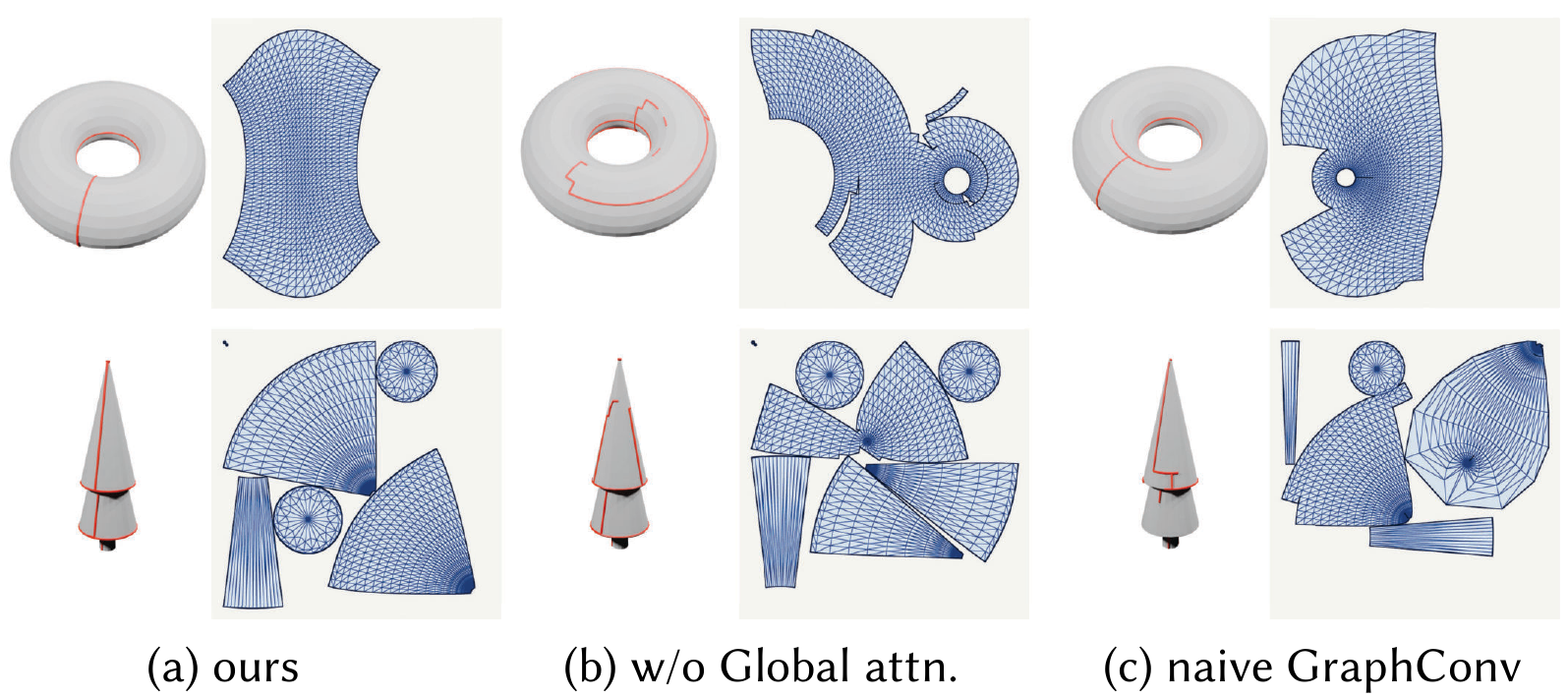}
    \caption{\textbf{Backbone design ablation.}
    Compared to the full model, removing global self-attention leads to less coordinated seam layouts and irregular UV partitions, while replacing local attention with a naive graph convolution produces less geometry-aware cuts and poorly organized UV islands.}
    \vspace{-6mm}
    \label{fig:ablation}
\end{figure}

As shown in Tab.~\ref{tab:structure}, we first evaluate the perceptual quality of the generated seam layouts using both automated and human assessments.
For the VLM-based evaluation, we concatenate anonymized UV layouts from all methods into a randomized panel for each test example and ask Gemini-2.5-Flash~\cite{gemini25} to assign a 1--5 quality score to each layout.
\methodname{} achieves the highest average VLM score, suggesting better structural clarity and paintability under this standardized evaluation protocol.
We further conduct a user study to assess human preference.
Twenty participants with 3D texturing experience rate each result on a 1--5 scale based on seam naturalness, chart coherence, and suitability for texture editing.
\methodname{} receives the highest average score, further confirming its alignment with artist-oriented UV editing criteria.
We then compare basic layout statistics, including the number of seam edges and the number of UV islands.
These measures reflect whether a method produces cuts and chart decompositions at a granularity similar to artist-authored layouts.
\methodname{} is closest to the ground truth on both measures, indicating that the learned generative prior captures the overall cutting style of professional UV layouts.
Finally, we report traditional geometric metrics, including distortion and runtime.
Interestingly, the ground-truth artist-authored layouts have the worst area distortion and the second-worst angular distortion among all methods, suggesting that distortion minimization is not the primary goal of professional seam placement.
Although \methodname{} does not explicitly optimize distortion objectives, it achieves angular distortion comparable to existing automatic unwrapping methods, while its area distortion remains within a reasonable range.
In terms of runtime, \methodname{} is slightly slower than Blender, xatlas, and PartUV, but much faster than OptCuts, providing a practical trade-off between artist-aligned seam quality and efficiency.

Further details, including  additional seam and island structure statistics, metric definitions, user-study examples, VLM prompts, and representative VLM evaluations, are provided in the supplementary material.

\subsection{Ablation Studies}
\label{sec:ablation}
We ablate the architectural design of our Mesh Transformer backbone by comparing \methodname{} with two variants: \emph{w/o Global Attn.}, which removes the global self-attention stream, and \emph{Naive GraphConv}, which replaces our local attention module with vanilla graph convolution while keeping the global stream. 
All variants are trained from scratch under the same settings and evaluated on the same 200-mesh validation set as Sec.~\ref{sec:quantitative}.
Since generated seams should lead to valid UV atlases, we report the overlap rate as a practical validity metric, measuring the fraction of meshes for which all four generated samples result in overlapping UV atlases.
As shown in the Tab~\ref{tab:backbone_ablation_structure}, the full model best matches the ground truth statistics, while both removing global self-attention and replacing local attention with naive graph convolution substantially increase the overlap rate. 
This confirms the importance of both global shape context and expressive local seam modeling for generating coherent and usable seam layouts.
Fig.~\ref{fig:ablation} provides qualitative evidence consistent with the quantitative results.
Without global self-attention, the model produces locally plausible but less coordinated seams, resulting in fragmented or irregular UV partitions.
With naive graph convolution, the predicted cuts become less aligned with salient geometric structures and yield poorly organized UV islands.
The full model produces cleaner seam paths and more regular UV islands.

\begin{table}[t]
    \centering
     \definecolor{1st}{HTML}{F8CBA6}
    \definecolor{2nd}{HTML}{FFE7CC}
    \caption{\textbf{Ablation on backbone design.}
   We compare the full Mesh Transformer with two variants: one without global self-attention and one replacing local attention with vanilla graph convolution.
The full model achieves the lowest overlap rate and best matches the ground-truth layout statistics, confirming the importance of both global context and expressive local seam modeling.
Ground-truth reference values are reported in Tab.~\ref{tab:structure}.}
    \label{tab:backbone_ablation_structure}
    \vspace{2pt}
    \small
    \setlength{\tabcolsep}{3.5pt}
    \begin{tabular}{lccc}
        \toprule
        Metric & Ours & w/o Global Attn & GraphConv \\
        \midrule
        \# Seam edges $\sim GT$ & \cellcolor{1st}471.5 &\cellcolor{2nd}448.5  & 261.9 \\
        \# Islands $\sim GT$& \cellcolor{1st}62.7  & \cellcolor{2nd} 61.1 & 29.3 \\
        Pred. Overlap Rate $\downarrow$ & \cellcolor{1st}{9\%} & \cellcolor{2nd}{24.5\%} & 52\% \\
        \bottomrule
    \end{tabular}
    \vspace{-2mm}
\end{table}

\begin{figure}[t]
    \centering
    \includegraphics[width=\columnwidth]{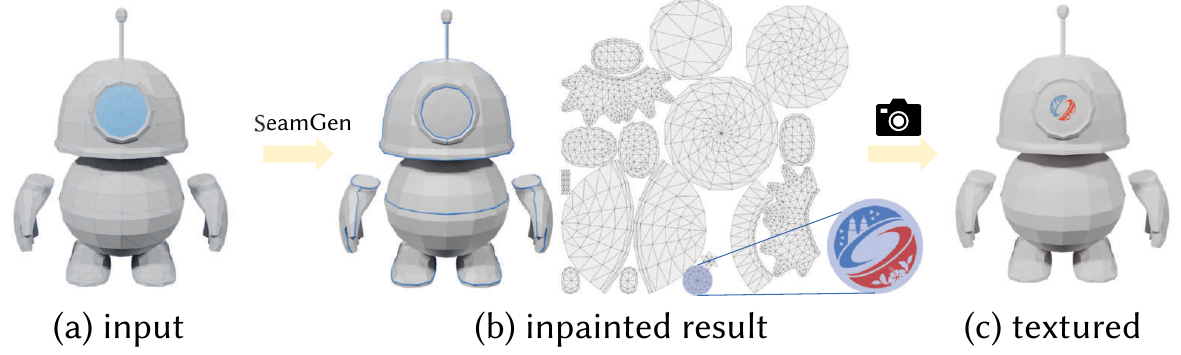}
    \caption{
    \textbf{User-guided seam generation via conditional inpainting.}
    (a) The user marks a seam-free region (blue) on the eye area of the mesh, specifying that this region should remain free of cuts.
    (b) Our method completes the remaining seam layout under this constraint and unwraps the mesh into a UV atlas.
    (c) The resulting seam-free eye region provides a clean, continuous surface for texture placement. We insert the SIGGRAPH Asia logo into the corresponding UV region and render the textured result.
    }
    \vspace{-4mm}
    \label{fig:user_guide_vis}
\end{figure}
\subsection{Additional Qualitative Results}
\label{sec:additional_results}

\begin{figure*}[p]
    \centering
    \includegraphics[width=0.96\linewidth]{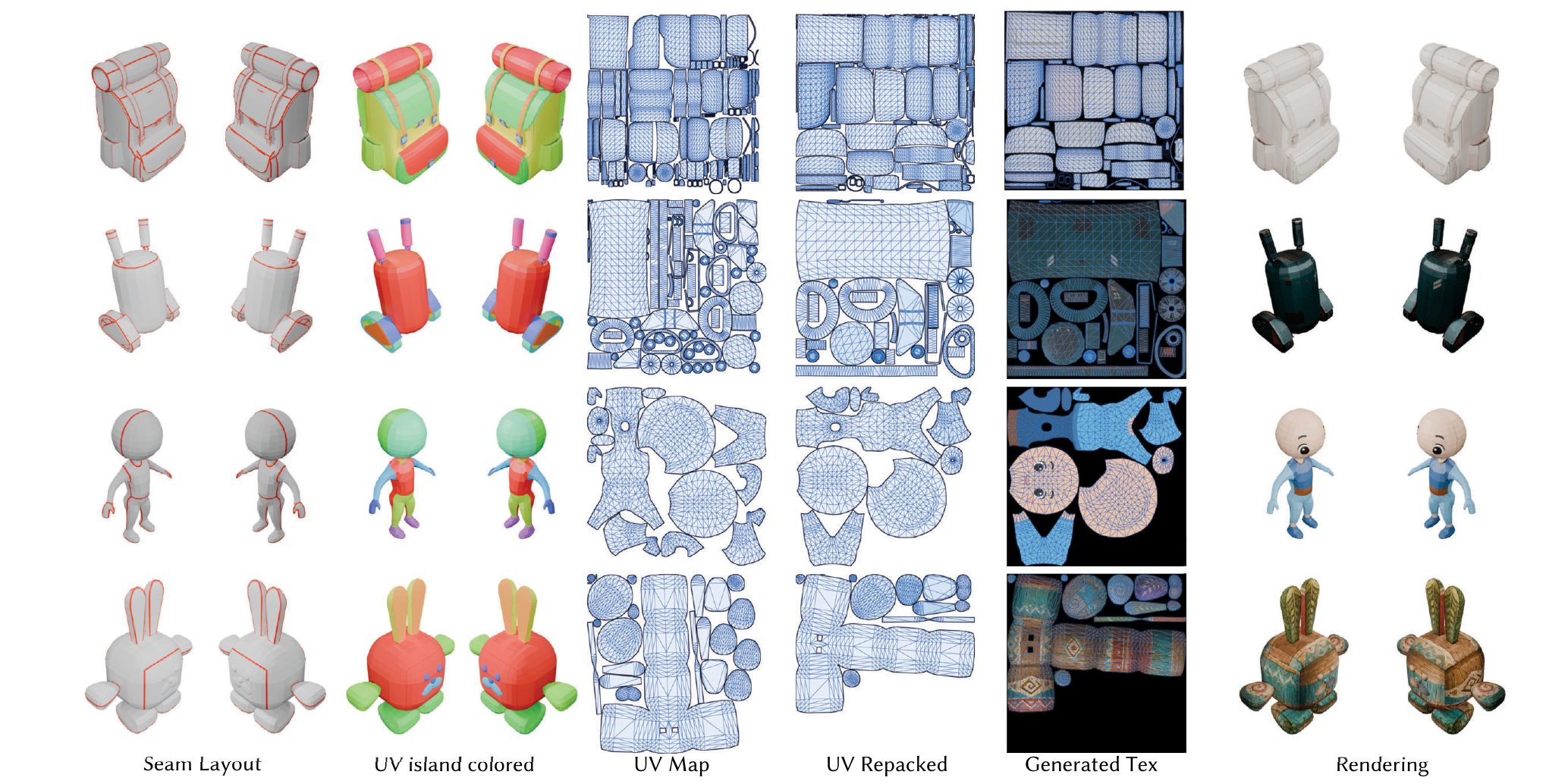}
    \caption{\textbf{Qualitative results of symmetric seam layouts and downstream texturing.} We show that our method inherently respects the structural symmetry of input meshes during seam generation, enabling symmetric parts to share the same UV space after repacking. This significantly enhances texture utilization, providing a superior foundation for downstream UV-space texture generation and editing.}
    \label{fig:sym}
\end{figure*}

\begin{figure*}[p]
    \centering
    \includegraphics[width=0.99\linewidth]{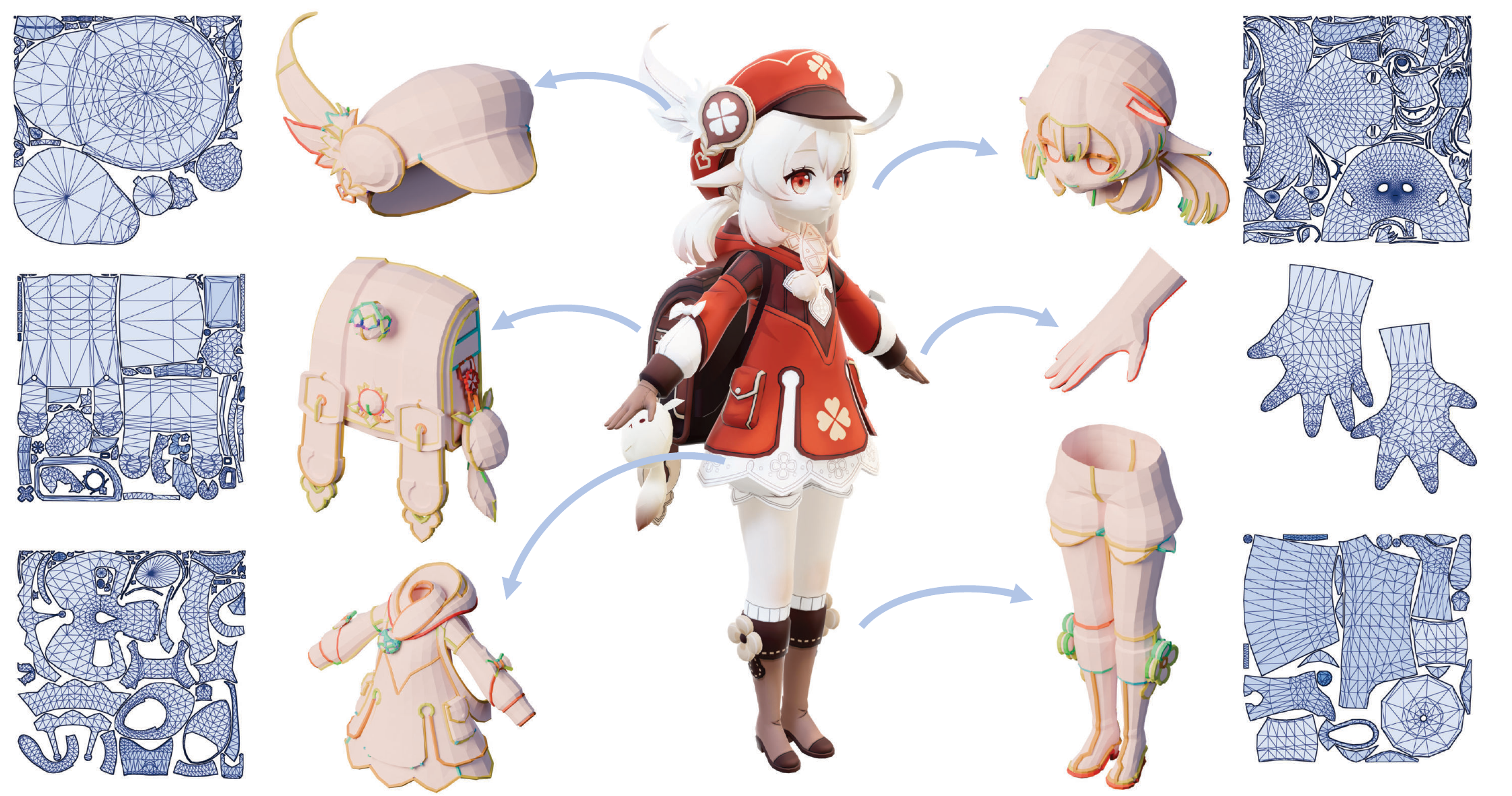}
    \caption{\textbf{Component-wise UV Seam Generation integrated with professional art workflows.} Although our model is primarily designed for meshes with moderate triangle counts, it perfectly adapts to professional production pipelines. In practice, artists typically decompose complex characters into multiple semantic parts (e.g., head, outfit, accessories) for UV unwrapping. Our method seamlessly integrates into this workflow, ensuring high-quality seam placement across all individual components.}
    \label{fig:part}
\end{figure*}

We further demonstrate two practical properties of \methodname{} in production-oriented UV workflows.
First, \methodname{} can preserves structural symmetry when generating seam layouts for symmetric objects. 
As shown in Fig.~\ref{fig:sym}, the generated seams are likely to be placed on corresponding symmetric parts, producing UV islands with compatible shapes after repacking. 
This makes it possible to reuse the same UV space across symmetric regions, improving texture-space utilization and providing a convenient basis for downstream UV-space texture generation and editing. 
We further demonstrate this benefit by generating image textures on the repacked UV layouts using GPT-Image-2~\cite{openai2026gptimage2}.
Second, we evaluate \methodname{} in a component-wise UV unwrapping workflow for complex production assets. 
Although our model is primarily designed for meshes with moderate triangle counts, it can be directly applied to complex models after they are decomposed into semantic components, as commonly done by artists in professional pipelines. 
Fig.~\ref{fig:part} shows an example game asset separated into multiple parts, such as the head, outfit, and accessories. 
Applying \methodname{} to each component produces coherent seam layouts across the full asset, demonstrating its compatibility with practical artist workflows.

\subsection{Conditional Inpainting Applications}
\label{sec:app_inpainting}
\begin{figure}[t]
    \centering
    \includegraphics[width=0.99\columnwidth]{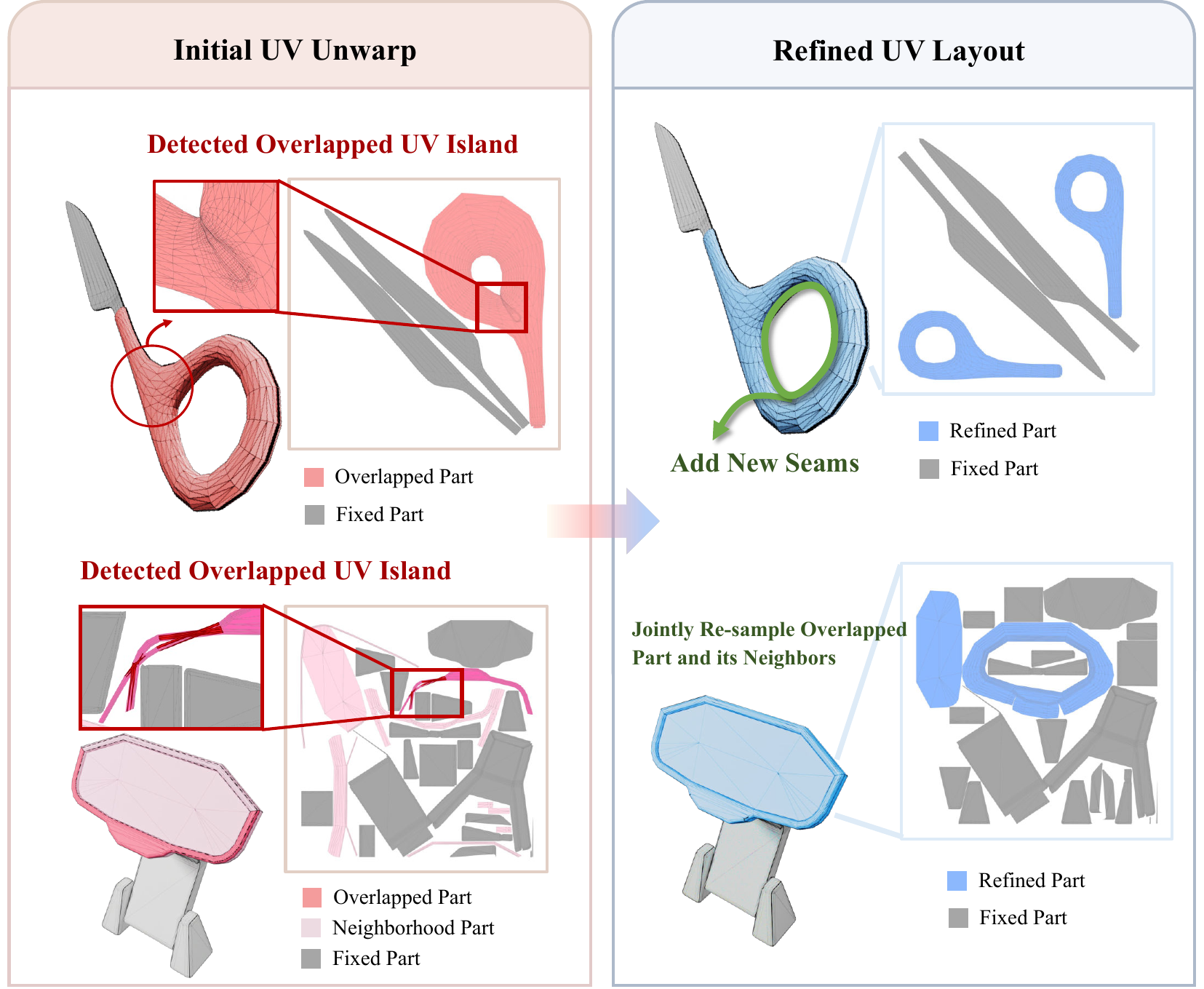}
    \vspace{-1mm}
    \caption{
    \textbf{Self-refinement via conditional seam inpainting.}
    Left: initial layouts with detected overlapping UV islands (red) and their neighborhood (pink). Right: refined, overlap-free UV islands (blue) with updated seam. Top: interior refinement adds new seams (green) within the problematic island. Bottom: boundary-aware refinement jointly resamples the overlapping island and its neighbors. 
    }
    \vspace{-2mm}
    \label{fig:refinement_vis}
\end{figure}

We demonstrate the practical utility of conditional seam refinement via the inpainting mechanism introduced in Sec.~\ref{sec:inpainting}.
First, Fig.~\ref{fig:user_guide_vis} demonstrates constraint-guided seam generation. 
Given a user-specified region where seams should be avoided, \methodname{} respects this constraint and completes the remaining cuts into a coherent seam layout. 
This provides simple local control while producing UV layouts suitable for subsequent texture painting.
Second, Fig.~\ref{fig:refinement_vis} shows that the same conditional sampling mechanism can be used for automatic self-refinement. 
When a sampled seam layout produces overlapping UV triangles, \methodname{} locally repairs the problematic region either by adding seams inside the overlapping island or by resampling its boundary together with neighboring islands. 
This corrects many invalid layouts without discarding the entire sample or training an additional repair model. 
Quantitatively, the single-sample predicted overlap rate is reduced from 26.5\% to 3.5\%, demonstrating improved atlas validity.
The detailed refinement procedure is provided in the supplementary material.

\section{Conclusion, Limitations, and Future Work}
We presented \methodname{}, a flow-matching generative framework that learns the distribution of artist-authored UV seam layouts directly on the mesh graph. By designing a Mesh Transformer backbone that interleaves local message passing along mesh edges with global self-attention over all vertices, \methodname{} captures both fine-grained geometric cues and long-range context required for production-quality seam placement. A training-free inpainting procedure inherited from the flow matching formulation enables both user-guided seam generation and iterative resolution of overlapping UV islands with a single pretrained model, without heuristic post-processing or auxiliary training objectives. Experiments show that \methodname{} produces seam layouts whose seam-structure and UV-island statistics closely match the ground-truth distribution, outperforming distortion-based and semantic-proxy baselines across quantitative metrics, qualitative comparisons, and user studies.

Despite these results, several limitations remain. 
First, evaluating whether a generated seam distribution closely matches the ground truth remains challenging. 
There is still no direct and widely accepted metric for measuring distribution-level alignment with artist-authored seam layouts. 
Although we use UV-layout statistics, VLM-based evaluation, and a user study as complementary evidence, designing better objective metrics for artist-aligned seam assessment remains an important direction for future work.
Second, \methodname{} may occasionally produce invalid UV atlases. 
In the worst case, a small number of problematic islands may require local fallback to traditional unwrapping methods.
Finally, our current implementation may exceed GPU memory for meshes with more than roughly 30K faces. 
Future hierarchical graph designs could improve scalability to high-resolution assets.


{
    \small
    \bibliographystyle{ieeenat_fullname}
    \bibliography{main}
}

\end{document}